\documentclass[12pt]{article}

\usepackage[margin=1in]{geometry}
\usepackage{amsmath,amssymb,amsthm}
\usepackage{graphicx}
\usepackage{booktabs}
\usepackage{multirow}
\usepackage{array}
\usepackage{tabularx}
\usepackage{hyperref}
\usepackage{natbib}
\usepackage{setspace}
\usepackage{float}
\usepackage{caption}
\usepackage{subcaption}
\usepackage{xcolor}
\usepackage{enumitem}
\usepackage[title]{appendix}
\usepackage{tikz}
\usetikzlibrary{arrows.meta, positioning, shapes.geometric, calc, fit, backgrounds}

\hypersetup{colorlinks=true, linkcolor=blue, citecolor=blue, urlcolor=blue}
\title{\textbf{Nonstandard Errors in AI Agents}}

\author{
Ruijiang Gao \qquad Steven Chong Xiao \\[4pt]
{\normalsize Naveen Jindal School of Management, University of Texas at Dallas} \\
{\normalsize \texttt{ruijiang.gao@utdallas.edu} \qquad \texttt{steven.xiao@utdallas.edu}}
}

\date{}

\begin{document}

\maketitle

\begin{abstract}\small
We study whether state-of-the-art AI coding agents, given the same data and research question, produce the same empirical results. Deploying 150 autonomous Claude Code agents to independently test six hypotheses about market quality trends in NYSE TAQ data for SPY (2015--2024), we find that AI agents exhibit sizable \textit{nonstandard errors} (NSEs), that is, uncertainty from agent-to-agent variation in analytical choices, analogous to those documented among human researchers. AI agents diverge substantially on measure choice (e.g., autocorrelation vs.\ variance ratio, dollar vs.\ share volume). Different model families (Sonnet 4.6 vs.\ Opus 4.6) exhibit stable ``empirical styles,'' reflecting systematic differences in methodological preferences. In a three-stage feedback protocol, AI peer review (written critiques) has minimal effect on dispersion, whereas exposure to top-rated exemplar papers reduces the interquartile range of estimates by 80--99\% within \textit{converging} measure families. Convergence occurs both through within-family estimation tightening and through agents switching measure families entirely, but convergence reflects imitation
rather than understanding. These findings have implications for the growing use of AI in automated policy evaluation and empirical research.

\end{abstract}

\section{Introduction}
\label{sec:intro}

The deployment of artificial intelligence in empirical research is accelerating rapidly. AI agents now automate tasks ranging from data collection and coding to statistical estimation and report writing \citep{lu2024aiscientist}. Projects such as Autonomous Policy Evaluation \citep[APE;][]{ape2026} deploy AI to produce end-to-end economics research papers, raising a fundamental question: if we ask different AI agents to analyze the same data and test the same hypothesis, will they produce the same result?

The concern is well motivated. A large literature documents that ``researcher degrees of freedom'' \citep{simmons2011false} introduce substantial variation into empirical findings. \citet{silberzahn2018many} show that 29 teams analyzing the same soccer data reach divergent conclusions about whether dark-skinned players receive more red cards. \citet{botvinik2020variability} find that 70 neuroimaging teams testing nine hypotheses on the same fMRI data produce highly variable results. \citet{menkveld2024nonstandard} document that 164 research teams, given the same data and hypotheses about market microstructure, produce estimates with an interquartile range (IQR) comparable in magnitude to the standard error of any individual estimate. They use the term \textit{nonstandard errors} (NSE) to describe this additional uncertainty arising from researcher-to-researcher variation in analytical choices in the ``evidence-generating process'' (EGP).

A natural question is whether AI agents, which lack the idiosyncratic training and institutional pressures that shape human researchers' choices, would exhibit similar variation. On one hand, AI agents drawing on the same foundation model might converge to a single ``best'' approach, eliminating NSE. On the other hand, the stochasticity inherent in language model sampling, combined with the underspecification of most empirical tasks \citep{steegen2016increasing}, could produce meaningful variation even among identically initialized agents.

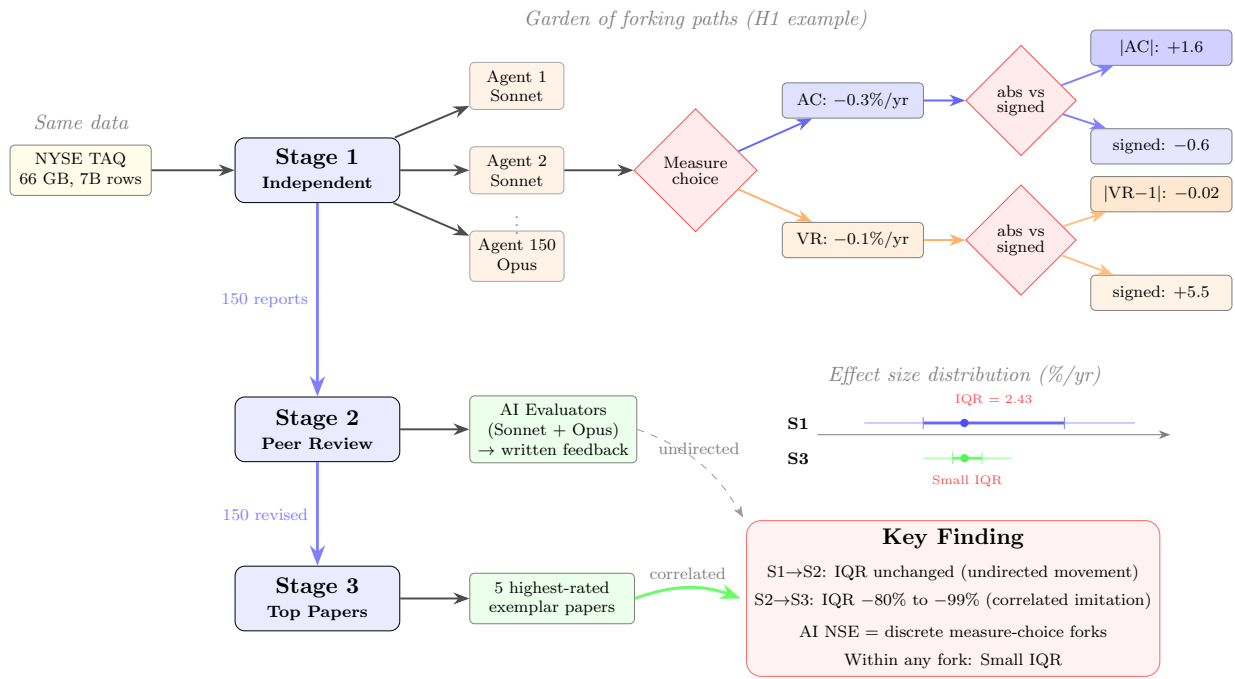
\begin{figure}[t]
\centering
\resizebox{\textwidth}{!}{
\begin{tikzpicture}[scale=0.85, every node/.style={scale=0.85},
    >=Stealth,
    node distance=1.2cm and 1.8cm,
    stage/.style={rectangle, rounded corners=4pt, draw=black, fill=blue!8,
                  minimum width=2.8cm, minimum height=1.1cm, align=center, font=\small\bfseries},
    agent/.style={rectangle, rounded corners=2pt, draw=gray!70, fill=orange!10,
                  minimum width=1.6cm, minimum height=0.6cm, align=center, font=\scriptsize},
    feedback/.style={rectangle, rounded corners=2pt, draw=black!50, fill=green!8,
                     minimum width=2.8cm, minimum height=0.8cm, align=center, font=\scriptsize},
    result/.style={rectangle, rounded corners=2pt, draw=black!50, fill=yellow!10,
                   minimum width=2.4cm, minimum height=0.6cm, align=center, font=\scriptsize},
    fork/.style={diamond, draw=red!70, fill=red!8, minimum width=1.2cm,
                 minimum height=0.8cm, align=center, font=\scriptsize},
    arr/.style={->, thick, color=black!70},
    darr/.style={->, dashed, color=black!40},
]

\node[stage] (s1) {Stage 1\\[-2pt]\scriptsize Independent};

\node[agent, above right=0.4cm and 1.0cm of s1] (a1) {Agent 1\\Sonnet};
\node[agent, right=1.0cm of s1] (a2) {Agent 2\\Sonnet};
\node[agent, below right=0.4cm and 1.0cm of s1] (a3) {Agent 150\\Opus};
\node[font=\scriptsize, color=gray] at ($(a2)!0.5!(a3) + (0, -0.1)$) {$\vdots$};

\node[result, left=1.2cm of s1] (data) {NYSE TAQ\\66 GB, 7B rows};
\draw[arr] (data) -- (s1);

\draw[arr] (s1) -- (a1);
\draw[arr] (s1) -- (a2);
\draw[arr] (s1) -- (a3);

\node[fork, right=1.0cm of a2] (f1) {\scriptsize Measure\\[-2pt]\scriptsize choice};
\draw[arr] (a2) -- (f1);

\node[result, above right=0.3cm and 0.8cm of f1, fill=blue!12] (r1a) {AC: $-0.3$\%/yr};
\node[result, below right=0.3cm and 0.8cm of f1, fill=orange!12] (r1b) {VR: $-0.1$\%/yr};
\draw[arr, color=blue!60] (f1) -- (r1a);
\draw[arr, color=orange!60] (f1) -- (r1b);

\node[fork, right=0.6cm of r1a] (f2) {\scriptsize abs vs\\[-2pt]\scriptsize signed};
\draw[arr, color=blue!60] (r1a) -- (f2);

\node[result, above right=0.1cm and 0.6cm of f2, fill=blue!18] (r2a) {\scriptsize $|$AC$|$: $+1.6$};
\node[result, below right=0.cm and 0.6cm of f2, fill=blue!10] (r2b) {\scriptsize signed: $-0.6$};
\draw[arr, color=blue!50] (f2) -- (r2a);
\draw[arr, color=blue!50] (f2) -- (r2b);

\node[fork, right=0.6cm of r1b] (f3) {\scriptsize abs vs\\[-2pt]\scriptsize signed};
\draw[arr, color=orange!60] (r1b) -- (f3);

\node[result, above right=0.cm and 0.6cm of f3, fill=orange!18] (r3a) {\scriptsize $|$VR$-1|$: $-0.02$};
\node[result, below right=0.1cm and 0.6cm of f3, fill=orange!10] (r3b) {\scriptsize signed: $+5.5$};
\draw[arr, color=orange!50] (f3) -- (r3a);
\draw[arr, color=orange!50] (f3) -- (r3b);

\node[stage, below=2.8cm of s1] (s2) {Stage 2\\[-2pt]\scriptsize Peer Review};
\node[feedback, right=1.0cm of s2] (fb) {AI Evaluators\\(Sonnet + Opus)\\$\to$ written feedback};
\draw[arr] (s2) -- (fb);
\draw[darr, bend left=20] (fb.east) to node[above, font=\scriptsize, color=gray] {undirected} ($(fb.east) + (1.8, -1.5)$);

\draw[arr, very thick, color=blue!50] (s1.south) -- node[left, font=\scriptsize] {150 reports} (s2.north);

\node[stage, below=1.5cm of s2] (s3) {Stage 3\\[-2pt]\scriptsize Top Papers};
\node[feedback, right=1.0cm of s3] (tp) {5 highest-rated\\exemplar papers};
\draw[arr] (s3) -- (tp);
\draw[arr, very thick, color=green!60, bend left=20] (tp.east) to node[above, font=\scriptsize, color=gray] {correlated} ($(tp.east) + (1.8, 0)$);

\draw[arr, very thick, color=blue!50] (s2.south) -- node[left, font=\scriptsize] {150 revised} (s3.north);

\node[rectangle, rounded corners=6pt, draw=red!60, fill=red!5,
      minimum width=5cm, align=center, font=\small,
      right=5.0cm of s3] (insight) {
  \textbf{Key Finding}\\[2pt]
  \scriptsize S1$\to$S2: IQR unchanged (undirected movement)\\
  \scriptsize S2$\to$S3: IQR $-$80\% to $-$99\% (correlated imitation)\\[2pt]
  \scriptsize AI NSE $=$ discrete measure-choice forks\\
  \scriptsize Within any fork: Small IQR
};

\node[font=\footnotesize\itshape, color=gray] (distlabel) at (11, -3.5) {Effect size distribution (\%/yr)};

\begin{scope}[shift={($(distlabel) + (0.0, -1.0)$)}]
  \draw[->, gray, thin] (-2.5, 0) -- (3.5, 0);
  \draw[very thick, blue!60] (-0.7, 0.2) -- (1.7, 0.2);
  \draw[thin, blue!40] (-1.7, 0.2) -- (-0.7, 0.2);
  \draw[thin, blue!40] (1.7, 0.2) -- (2.9, 0.2);
  \fill[blue!70] (0.0, 0.2) circle (2pt);
  \draw[blue!50] (-0.7, 0.1) -- (-0.7, 0.3);
  \draw[blue!50] (1.7, 0.1) -- (1.7, 0.3);
  \node[font=\scriptsize\bfseries, left] at (-2.5, 0.2) {S1};
  \node[font=\tiny, color=red!70, above] at (0.5, 0.35) {IQR = 2.43};

  \draw[very thick, green!60] (-0.2, -0.4) -- (0.3, -0.4);
  \draw[thin, green!40] (-0.7, -0.4) -- (-0.2, -0.4);
  \draw[thin, green!40] (0.3, -0.4) -- (0.8, -0.4);
  \fill[green!70] (0.0, -0.4) circle (2pt);
  \draw[green!50] (-0.2, -0.5) -- (-0.2, -0.3);
  \draw[green!50] (0.3, -0.5) -- (0.3, -0.3);
  \node[font=\scriptsize\bfseries, left] at (-2.5, -0.4) {S3};
  \node[font=\tiny, color=red!70, below] at (0.05, -0.55) {Small IQR};
\end{scope}

\node[font=\footnotesize\itshape, color=gray, above=0.1cm of data] {Same data};
\node[font=\footnotesize\itshape, color=gray, above=1cm of f1] {Garden of forking paths (H1 example)};

\end{tikzpicture}
}
\caption{Experimental design overview. 150 AI agents independently analyze the same NYSE TAQ data (Stage 1), receive AI peer review (Stage 2), and see the five highest-rated exemplar papers (Stage 3). At each stage, agents face a ``garden of forking paths'' \citep{steegen2016increasing}: measure choice, functional form, and other decision forks. Peer review causes undirected movement (IQR unchanged); exemplar exposure causes correlated imitation (IQR collapses within measure families).}
\label{fig:overview}
\end{figure}

We address this question by replicating the \citet{menkveld2024nonstandard} experimental design with AI agents. We deploy 150 autonomous Claude Code agents (100 Sonnet 4.6, 50 Opus 4.6) to independently test six hypotheses about market quality trends in SPY using NYSE TAQ millisecond data (2015--2024). Each agent operates as a fully autonomous researcher: it reads instructions, explores the data, constructs measures, estimates trends, writes a 2,000--4,000-word research report, and produces code, figures, and structured results. The entire pipeline, from data to deliverables, runs fully autonomously without human intervention.
We conduct a three-stage feedback protocol that mimics a fully automated AI research society: Stage 1 (independent analysis), Stage 2 (AI peer review with written feedback), and Stage 3 (exposure to the five highest-rated exemplar papers). We analyze the dynamics of AI NSE in these stages. 
The complete workflow is shown in Figure \ref{fig:overview}.

First, we find that AI agents exhibit sizable nonstandard errors. 
The effect can range from negative to positive, insignificant to significant, and small to large. For example, for the market efficiency hypothesis (H1), the IQR of effect size estimates across agents is 2.43\%/yr, with estimates ranging from $-$0.74\%/yr to +1.7\%/yr.
For the hypothesis on trading volume (H4), the IQR of effect size estimates across agents is 10.69\%/yr, driven entirely by agents choosing different volume measures (dollar volume at approximately +6\%/yr vs.\ share volume at approximately $-$5\%/yr). For price impact (H6), the IQR is 10.34\%/yr, reflecting a split between trade-level price impact and Amihud illiquidity ratio measures \citep{amihud2002illiquidity}.

Second, the structure of AI NSE is distinctive. All 150 agents independently choose regression with a linear time trend as the estimation paradigm, split between level specifications (56\%) and log specifications (44\%). No agent uses relative changes (period-over-period ratios), first differences, that are common in human choices. AI NSE is therefore concentrated almost entirely in measure-choice forks (which measure to compute) rather than in estimation-paradigm forks (how to estimate a given measure). Notably, the relative-change specification, which \citet{menkveld2024nonstandard} identify as a major source of dispersion through Jensen's inequality bias, is entirely absent from the AI agents' repertoire.

Third, different model families exhibit stable ``empirical styles.'' Sonnet 4.6 agents overwhelmingly prefer autocorrelation measures for market efficiency (87\%), level OLS, and daily frequency. Opus 4.6 agents universally choose variance ratio measures (100\%), prefer log OLS, and use monthly frequency 28\% of the time. These preferences are systematic rather than random, consistent with recent evidence that LLMs embed stable biases in code generation \citep{zhang2025invisible}.

Fourth, we implement a three-stage feedback protocol consisting of independent analysis (S1), AI peer review (S2), and exposure to the five highest-rated papers (S3), analogous to stages S1--S3 in \citet{menkveld2024nonstandard}. We find that AI peer review has no effect on reducing dispersion: the IQR is essentially unchanged from S1 to S2. In contrast, exposure to top papers triggers dramatic convergence. For converging measure families, the IQR collapses by 80--99\% from S1 to S3. However, for two hypotheses (H1 and H5), dispersion increases as exemplar papers introduce new methodological options that agents adopt inconsistently. A striking additional mechanism is \textit{measure-family migration}: 62 of 87 autocorrelation agents switch to variance ratio after seeing top papers that use variance ratio measures. 

Our paper contributes to several literatures. We add to the growing body of work on replication and the credibility of empirical research \citep{simmons2011false,harvey2016and,brodeur2020methods}. We extend the NSE framework of \citet{menkveld2024nonstandard} from human researchers to AI agents, showing that the problem of analytical variation persists even when human judgment is removed from the pipeline. We contribute to the literature on multiverse analysis \citep{steegen2016increasing} by providing a large-scale demonstration that AI-generated multiverses may be structurally narrower than human-generated ones. Finally, we inform the nascent policy discussion around automated research by documenting the NSE in AI agents in empirical analysis.

The remainder of this paper is organized as follows. Section~\ref{sec:related} reviews related work. Section~\ref{sec:design} describes the experimental design. Section~\ref{sec:nse} documents the existence and structure of AI NSE. Section~\ref{sec:feedback} analyzes the effects of peer feedback and exemplar exposure. Section~\ref{sec:multiverse} presents the multiverse analysis of decision forks. Section~\ref{sec:discussion} discusses implications. Section~\ref{sec:conclusion} concludes.

\section{Related Work}
\label{sec:related}

\subsection{Automated Research with AI Agents}

AI systems are increasingly deployed not merely as tools for researchers but as autonomous researchers themselves. \citet{lu2024aiscientist} introduce the AI Scientist, a system that autonomously generates hypotheses, designs experiments, runs code, and writes full scientific papers at a cost of less than \$15 per paper. Project APE \citep{ape2026} deploys AI agents for end-to-end economics policy evaluation. These systems raise a fundamental question that existing work has not addressed: \textit{how reliable are the empirical estimates they produce?} A single demonstration that an AI agent can write a paper tells us nothing about whether a different run, or a different model, would reach the same conclusion.

In concurrent and independent work, \citet{huang2026ai} run 158 instances of GPT-5.2 on the \#fincap dataset from \citet{menkveld2024nonstandard} and find that AI agents concentrate on a narrow set of analysis paths. Our study differs in three important ways. First, our agents are fully autonomous tool-using systems (Claude Code) that interactively explore data, write and debug code, and produce complete research reports, whereas their agents are customized themselves and generate code in a single shot against dataset metadata without seeing the data. This distinction matters because interactive data exploration may lead to different methodological choices than blind code generation. Second, we implement a three-stage feedback protocol (independent analysis, AI peer review, exemplar exposure) that allows us to study how AI agents respond to social scientific feedback mechanisms. Third, we deploy two model families (Sonnet and Opus) within the same architecture, enabling us to decompose AI NSE into within-model stochasticity and between-model ``empirical style'' components.

\subsection{Nonstandard Errors in Social Science}

A growing literature documents that researcher-to-researcher variation in analytical choices, termed ``researcher degrees of freedom'' by \citet{simmons2011false}, introduces substantial uncertainty into empirical findings. This literature has progressed through increasingly large-scale crowdsourced experiments. \citet{silberzahn2018many} recruit 29 teams to analyze the same soccer data and find conclusions ranging from a significant positive to a significant negative effect. \citet{botvinik2020variability} scale to 70 neuroimaging teams testing nine hypotheses on the same fMRI data. \citet{breznau2022observing} study 73 teams analyzing identical cross-country survey data on immigration attitudes, finding that observable team characteristics explain only 4\% of the variability in results, with the remaining 96\% attributable to undocumented analytical choices.

\citet{menkveld2024nonstandard} formalize this phenomenon as ``nonstandard errors'' (NSE) and conduct the largest such study to date, with 164 research teams and 34 peer evaluators testing six hypotheses on EURO STOXX 50 futures data. They introduce a multiverse-based diagnostic that identifies sampling frequency and model choice as the dominant sources of dispersion, and show that peer feedback reduces NSE by 47\% across four stages. The multiverse analysis framework of \citet{steegen2016increasing} provides a complementary single-researcher approach, while \citet{brodeur2020methods} document the downstream consequences of analytical flexibility at scale.

A critical gap in this literature is that all existing many-analyst studies use \textit{human} researchers, leaving open whether the observed variation reflects irreducible task ambiguity or idiosyncratic human factors (training, incentives, fatigue). Our study addresses this gap by replacing human researchers with AI agents, which share training data and architecture, thereby isolating the variation attributable to task underspecification from that attributable to researcher heterogeneity. If AI agents, which lack idiosyncratic human biases, still exhibit substantial NSE, the variation must arise from the research question itself rather than from the researchers.

\subsection{LLM Behavioral Research}

A parallel literature investigates the behavioral properties of LLMs as computational models of human decision-making. \citet{horton2023homo} propose treating LLMs as ``Homo silicus'': simulated economic agents that can be endowed with preferences and information and whose behavior can be explored via simulation. \citet{argyle2023out} demonstrate ``algorithmic fidelity,'' showing that GPT-3, when conditioned on sociodemographic backstories, generates response distributions that closely match real survey data from diverse human subpopulations.

This work establishes that LLMs encode rich models of human behavior, but it also reveals systematic biases. \citet{perez2023discovering} and \citet{sharma2024towards} document \textit{sycophancy}: the tendency of LLMs to conform to a user's stated opinion even when incorrect, a bias arising from reinforcement learning from human feedback. Sycophancy has been studied primarily in conversational settings, but our findings suggest it has a direct analog in research methodology: when AI agents see exemplar papers in Stage 3, they engage in wholesale imitation rather than independent evaluation, switching measures to match the exemplar regardless of whether the switch is analytically justified (Section~\ref{sec:feedback}).

Our work bridges and extends these three literatures. Unlike the automated research literature, we do not ask whether AI can \textit{do} research but whether AI research outputs are \textit{reliable} across runs. Unlike the NSE literature, we use AI agents as a controlled probe to separate task-inherent ambiguity from researcher-specific variation. Unlike the LLM behavioral literature, we study LLM behavior not through surveys or games but through the complete pipeline of empirical research, from data exploration to report writing. The result is a new lens on the credibility of AI-generated empirical evidence, with implications for both the deployment of automated research systems and the interpretation of the human NSE literature.

\section{Experimental Design}
\label{sec:design}

\subsection{The \#AIcap Project}

We design the \#AIcap project as an AI analog of the \#fincap crowdsourced research project in \citet{menkveld2024nonstandard}. Where \#fincap recruited 164 human research teams, we deploy 150 AI coding agents. Each agent receives identical instructions and access to the same dataset, and independently produces a research report with quantitative estimates.

\paragraph{AI agents.} Each agent is an instance of Claude Code (Anthropic), a command-line AI coding agent that can read files, write code, execute programs, inspect outputs, and iteratively refine its analysis. We use two model variants: 100 agents run Sonnet 4.6 (\texttt{claude-sonnet-4-6}) and 50 agents run Opus 4.6 (\texttt{claude-opus-4-6}). Both models use default sampling parameters (temperature = 1.0). The agents operate with the \texttt{--dangerously-skip-permissions} flag, granting full autonomy over their workspace. Each agent has a budget cap of \$20, 40 GB RAM, and 6 CPUs.

\paragraph{Data.} The dataset comprises NYSE TAQ millisecond trade and quote data for the SPDR S\&P 500 ETF Trust (SPY) from January 2, 2015, through December 31, 2024. The data spans 2,516 trading days, approximately 66 GB, and over 7 billion rows. Agents access the data as read-only Parquet files mounted in a Singularity container.

\paragraph{Hypotheses.} Agents test six hypotheses about market quality trends over the sample period (2015--2024), each with the null of no change. The following descriptions are given verbatim to all agents (full instructions in Appendix~\ref{app:instructions}):
\begin{itemize}[nosep]
    \item[H1:] ``Assuming that informationally-efficient prices follow a random walk, did market efficiency change over time?''
    \item[H2:] ``The quoted bid-ask spread is the difference between the best ask and the best bid price. It is a standard measure of trading cost. Did the quoted bid-ask spread change over time?''
    \item[H3:] ``The realized spread could be thought of as the gross-profit component of the spread as earned by the liquidity provider. It compares the trade price to the midpoint some time after the trade. Did the realized bid-ask spread change over time?''
    \item[H4:] ``Did daily trading volume change over time?''
    \item[H5:] ``Intraday price volatility captures the magnitude of price fluctuations within a trading day. Did intraday volatility change over time?''
    \item[H6:] ``Price impact measures how much prices move in response to trading activity. It captures the information content of trades and the depth of the market. Did the price impact of trades change over time?''
\end{itemize}
Note that H2 and H3 provide explicit definitions of the measure, while H1 (``market efficiency''), H4 (``daily trading volume''), and H6 (``price impact'') leave the specific operationalization to the agent. We refer to H1, H4, and H6 as \textit{abstract hypotheses} because they describe the concept at a level of abstraction that admits multiple valid measures. This difference in specificity directly shapes the NSE results (Section~\ref{sec:nse}).

\paragraph{Deliverables.} Each agent produces: (i) a structured CSV with effect size, standard error, $t$-statistic, and direction for each hypothesis; (ii) a research report of 2,000--4,000 words; (iii) all analysis code with reproduction instructions; and (iv) figures.

\paragraph{Isolation.} Agents run inside Singularity containers with filesystem isolation. They cannot communicate with each other,  or see other agents' outputs. Variation across agents arises solely from the stochasticity of the language model's sampling process.
This is a critical design choice as during trial run, we observe agents can peak at each other's outputs when running without isolation, as coding agents have access to the filesystem and can read other agents' reports.

\subsection{Three-Stage Feedback Protocol}

Following \citet{menkveld2024nonstandard}, we implement a staged feedback protocol:

\begin{description}[nosep]
    \item[Stage 1: Independent analysis.] Each agent independently analyzes the data and produces its deliverables. This is analogous to a researcher working alone before receiving any feedback.
    \item[Stage 2: Peer review.] Each agent receives anonymized written evaluations from two AI peer evaluators (one Sonnet, one Opus) rating its Stage 1 report on a 0--10 scale per hypothesis, with detailed written feedback. The agent then revises its report and results. This is analogous to receiving colleague feedback or a referee report.
    \item[Stage 3: Top papers.] Each agent receives the five highest-rated anonymized Stage 2 reports (selected by average peer evaluation score) and may revise its analysis one final time. This is analogous to seeing competitive papers at a conference.
\end{description}

\citet{menkveld2024nonstandard} implement a fourth stage in which teams report a final ``Bayesian update'' estimate without necessarily producing new code. We omit this stage as the AI agents' final edits are finished.

\paragraph{Cost.} The total API cost for the experiment is \$1,558 across 450 agent-stage runs, with a median cost of \$3.17 per agent for Stage 1, \$1.90 for Stage 2, and \$2.87 for Stage 3. Opus agents cost approximately 1.6$\times$ more than Sonnet agents per stage. The median wall-clock time is 53 minutes for S1, 27 minutes for S2, and 25 minutes for S3. Appendix~\ref{app:instructions} reports the full cost and time distributions. By comparison, \citet{menkveld2024nonstandard} estimate approximately 27 person-years of effort from their 164 human teams, which would cost approximately \$2.7 million at a \$100,000 annual salary.

\subsection{Effect Size Normalization}

A methodological challenge specific to AI agents is that different agents report effect sizes in different units. Level-OLS agents report the raw change in the measure per year ($\Delta y / \text{yr}$), while log-OLS agents report $\beta \times 100 \approx$ percentage change per year. They are not directly comparable. 

We develop a conversion pipeline to normalize all effect sizes to a common unit: percentage change per year (\%/yr) relative to the mean level of the measure. For each agent, an Opus conversion agent reads the research agent's estimation code and processed data to: (i) classify the model as log, or, level; (ii) detect any pre-scaling of the slope (e.g., $\times 100$); (iii) compute the time-series mean of the dependent variable from the agent's own processed data; and (iv) convert level-model effect sizes via $\text{effect\_size\_pctyr} = (\text{effect\_size} / \text{mean}_y) \times 100$.


\section{AI Nonstandard Errors Exist and Are Structured}
\label{sec:nse}

\subsection{NSE Is Sizable}

Table~\ref{tab:nse_s1} reports the distribution of Stage 1 effect size estimates (\%/yr) across 150 agents for each hypothesis. NSE, measured as the IQR, ranges from 0.43 (H2, quoted spread) to 10.70 (H4, volume).

\begin{table}[H]
\centering
\caption{Stage 1 effect size distribution (\%/yr) across 150 AI agents}
\label{tab:nse_s1}
\small
\begin{tabular}{lrrrrrrrrr}
\toprule
Hyp & $N$ & Q10 & Q25 & Median & Q75 & Q90 & IQR & IDR \\
\midrule
H1 & 150 & $-$1.65 & $-$0.74 & $-$0.06 & 1.70 & 2.86 & 2.43 & 4.51 \\
H2 & 150 & $-$6.74 & $-$6.63 & $-$6.21 & $-$6.21 & $-$6.19 & 0.43 & 0.55 \\
H3 & 150 & 1.17 & 2.33 & 4.75 & 7.61 & 9.21 & 5.28 & 8.05 \\
H4 & 150 & $-$4.71 & $-$4.60 & 5.84 & 6.09 & 6.09 & 10.69 & 10.80 \\
H5 & 150 & 2.98 & 3.00 & 3.08 & 3.54 & 3.67 & 0.54 & 0.69 \\
H6 & 150 & $-$14.97 & $-$13.23 & $-$10.31 & $-$2.89 & $-$1.35 & 10.34 & 13.62 \\
\bottomrule
\end{tabular}
\begin{minipage}{\textwidth}
\vspace{4pt}
\footnotesize\textit{Notes:} Effect sizes are expressed as percentage change per year (\%/yr) relative to the measure's mean level. IQR $=$ Q75 $-$ Q25; IDR $=$ Q90 $-$ Q10. All estimates are from Stage 1 (independent analysis, before any feedback).
\end{minipage}
\end{table}

\begin{figure}[H]
\centering
\includegraphics[width=\textwidth]{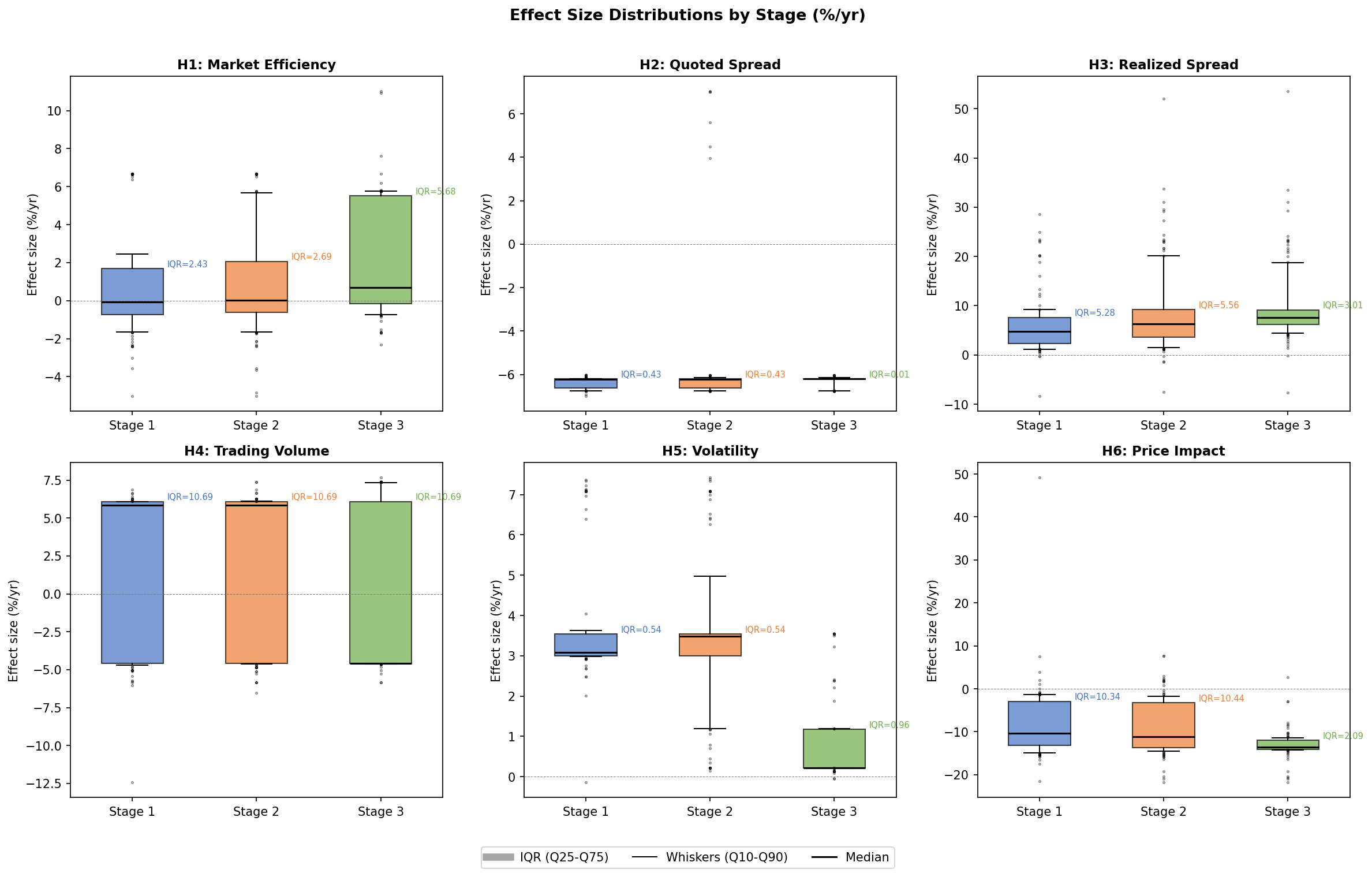}
\caption{Effect size distributions (\%/yr) by stage. Boxes show IQR (Q25--Q75) with IQR values annotated; whiskers extend to Q10 and Q90; horizontal line is the median. H2 and H5 have small IQR in S1, driven almost entirely by the log-vs-level specification fork (within either specification, agents agree to within 0.01\%/yr). H4 is bimodal in S1 (dollar vs.\ share volume). H5 collapses in S3 as 96\% of agents adopt year dummies. H6 converges in S3 as 99\% of agents adopt trade-level price impact.}
\label{fig:effect_distributions}
\end{figure}

Figure~\ref{fig:effect_distributions} displays the full distributions as violin plots with quantile markers.

The estimates reveal a clear pattern of heterogeneity. For H2 (quoted spread), agents are near-unanimous: all 150 find that spreads decreased, with a median of $-6.21$\%/yr and an IQR of only 0.43. In contrast, for H4 (volume), the distribution is sharply bimodal in S1: 90 agents using dollar volume find approximately $+6.1$\%/yr growth, while 60 agents using share volume find approximately $-4.6$\%/yr decline. (By S3, a third group of 17 agents uses number of trades, making the distribution trimodal.) The IQR of 10.70 is entirely the gap between these two measure families.

\subsection{AI NSE Is Driven by Measure Choice}

A key structural feature of AI NSE is its concentration in discrete measure-choice forks. Table~\ref{tab:nse_stratified} reports the IQR after stratifying H1 and H4 by measure family.

\begin{table}[H]
\centering
\caption{Stratified Stage 1 NSE: H1 and H4 by measure family}
\label{tab:nse_stratified}
\small
\begin{tabular}{lrrrr}
\toprule
Hypothesis & $N$ & Median & IQR & IDR \\
\midrule
H1:autocorrelation & 87 & $-$0.33 & 2.73 & 7.40 \\
H1:variance\_ratio & 63 & $-$0.05 & 1.45 & 2.70 \\
\midrule
H2 (all) & 150 & $-$6.21 & 0.43 & 0.55 \\
H3 (all) & 150 & 4.75 & 5.28 & 8.05 \\
\midrule
H4:dollar\_volume & 90 & 6.09 & 0.25 & 0.82 \\
H4:share\_volume & 60 & $-$4.60 & 0.11 & 0.55 \\
\midrule
H5 (all) & 150 & 3.08 & 0.54 & 0.69 \\
\midrule
H6:price\_impact & 84 & $-$13.03 & 2.59 & 5.33 \\
H6:amihud & 60 & $-$2.75 & 1.35 & 5.65 \\
H6:kyle\_$\lambda$ & 6 & 3.00 & 6.98 & --- \\
\bottomrule
\end{tabular}
\begin{minipage}{\textwidth}
\vspace{4pt}
\footnotesize\textit{Notes:} For H1 and H4, agents are stratified by measure family before computing the IQR. The unstratified H4 IQR of 10.69 reflects the gap between dollar volume ($+$6.1\%/yr) and share volume ($-$4.6\%/yr) families. Within each family, agents agree to within 0.25 and 0.11\%/yr, respectively.
\end{minipage}
\end{table}

We decompose AI NSE into three distinct sources, which vary in importance across hypotheses:

\begin{enumerate}[nosep]
\item \textbf{Hypothesis abstraction}: the research question is stated at a level of abstraction that admits multiple valid operationalizations (e.g., ``market efficiency'' can be measured via autocorrelation or variance ratio; ``trading volume'' can mean dollar or share volume; ``price impact'' can be trade-level or Amihud). This affects H1, H4, and H6.
\item \textbf{Specification-choice variation}: agents agree on the construct and measure but differ in functional form (log vs.\ level regression), frequency (daily vs.\ monthly), or estimation details (SE correction, weighting). This affects H2, H3, and H5.
\end{enumerate}

Table~\ref{tab:nse_decomp} summarizes the dominant source and within-source IQR for each hypothesis.

\begin{table}[H]
\centering
\caption{Sources of AI NSE by hypothesis (Stage 1)}
\label{tab:nse_decomp}
\footnotesize
\resizebox{\textwidth}{!}{
\begin{tabular}{llrp{5.5cm}}
\toprule
Hyp & Overall IQR & Dominant source & Within-source IQR \\
\midrule
H1 & 2.43 & Hypothesis abstraction (AC vs VR) + sub-forks (abs/signed, 1min/5min) & AC: 2.74, VR: 1.45. Within AC sub-cells: 0.02--0.34.\\
H2 & 0.43 & Specification (log vs level OLS) & Level: 0.003. Log: 0.11 \\
H3 & 5.28 & Specification (trade weighting) & Equal-wtd: 3.26. Vol-wtd: 2.04. Dollar-wtd: 1.26 \\
H4 & 10.69 & Hypothesis abstraction (dollar vs share volume) & Dollar: 0.25. Share: 0.11 \\
H5 & 0.54 & Specification (log vs level OLS) & Level: 0.005. Log: 0.12 \\
H6 & 10.34 & Hypothesis abstraction (price impact vs Amihud vs Kyle $\lambda$) & Price impact: 2.59. Amihud: 1.35 \\
\bottomrule
\end{tabular}
}
\begin{minipage}{\textwidth}
\vspace{4pt}
\footnotesize\textit{Notes:} ``Hypothesis abstraction'' (H1, H4, H6) means the instructions describe the hypothesis at a level of abstraction that admits multiple valid operationalizations. ``Specification choice'' (H2, H3, H5) means the agents agree on the measure but differ on functional form or estimation details.
\end{minipage}
\end{table}

The decomposition reveals a hierarchy. For the three abstract hypotheses (H1, H4, H6), the dominant source of NSE is hypothesis abstraction, which traces directly to the research instructions. The H1 instruction asks whether ``market efficiency'' changed without specifying which measure (autocorrelation or variance ratio), which return interval, or whether to use absolute or signed deviations. The H4 instruction asks whether ``daily trading volume'' changed without specifying dollar or share volume. The H6 instruction describes price impact as ``how much prices move in response to trading activity'' without specifying a particular measure. By contrast, the H2 instruction explicitly defines the quoted spread as ``the difference between the best ask and the best bid price,'' leaving little room for interpretation.

For H1, autocorrelation and variance ratio both measure market efficiency but capture different lag structures \citep[see][]{lo1988stock}, and within autocorrelation, the abs-vs-signed and 1min-vs-5min sub-forks further split agents into four clusters with internal IQRs ranging from 0.02 to 0.34. For H4, dollar volume and share volume measure fundamentally different things (capital flows vs.\ physical trading activity), and SPY's price approximately doubled over the sample period (2015--2024), mechanically causing these trends to diverge. For H6, trade-level price impact and the \citet{amihud2002illiquidity} ratio capture different aspects of market depth. Within each interpretation, agents agree tightly (IQR $\leq 0.25$\%/yr for H4, $\leq 2.59$ for H6, and $\leq 1.45$ for H1).

For H2 and H5 (well-specified hypotheses), hypothesis abstraction is absent and measure choice is not a factor. The entire IQR reflects specification choice: whether to use log or level regression. Within either specification, agents agree to within 0.003--0.005\%/yr. H3 falls between these groups: the measure (realized spread) is specified, but the aggregation method (equal vs.\ volume-weighted) is left to the agent.

The economic plausibility of the effect sizes can be verified against the raw data. For H2, the median estimate of $-6.21$\%/yr corresponds to SPY's time-weighted quoted spread declining from approximately 0.49 bps in 2015 to 0.28 bps in 2024, a 43\% decline consistent with a $-6.1$\%/yr geometric annual rate.

\subsection{Paradigm Uniformity}

All 150 agents independently choose to estimate trends by regressing the measure on a time trend via OLS. The variation is confined to:
\begin{itemize}[nosep]
    \item Functional form: level OLS (56\%) vs.\ log OLS (44\%)
    \item Standard error correction: HAC/Newey-West (80\%), robust/HC3 (11\%), none (9\%)
    \item Frequency: daily (86\%), monthly (13\%), annual (1 agent)
\end{itemize}

This contrasts with human teams in \citet{menkveld2024nonstandard}, where 35\% use regression with a linear time trend, 58\% use relative changes (period-over-period ratios), and 6\% use log-differences. Both human and AI researchers employ linear trend regression and log specifications, but the relative-change paradigm, which is the most common among humans, is entirely absent among AI agents.

One contributing factor may be the instructions themselves. Agents are asked to ``estimate the average per-year change in percentage points'' and shown an example reporting a linear annual decline with a standard error. This phrasing may elicit OLS regression of the measure on a time trend. Human researchers, drawing on broader methodological training, interpret similar instructions more flexibly. However, there is no explicit prompt to use OLS, or any specific estimation methods at all in the provided instructions, and the agents are free to deviate from the OLS paradigm if they choose.

\subsection{Model-Specific Empirical Styles}

Different model families exhibit systematic methodological preferences that we term ``empirical styles.'' Table~\ref{tab:model_prefs} reports key fork choices by model.

\begin{table}[H]
\centering
\caption{Model-specific empirical styles (Stage 1)}
\label{tab:model_prefs}
\footnotesize
\begin{tabular}{llrrl}
\toprule
Hypothesis & Fork & Sonnet 4.6 ($N$=100) & Opus 4.6 ($N$=50) & Test \\
\midrule
\multirow{4}{*}{H1} & Autocorrelation & 87\% & 0\% & \multirow{4}{*}{10.97***} \\
 & Variance ratio & 13\% & 100\% & \\
 & Level OLS & 96\% & 36\% & \\
 & Log OLS & 4\% & 64\% & \\
\midrule
\multirow{2}{*}{H2} & Level OLS & 90\% & 14\% & \multirow{2}{*}{37.64***} \\
 & Log OLS & 10\% & 86\% & \\
\midrule
\multirow{4}{*}{H3} & Level OLS & 99\% & 62\% & \multirow{4}{*}{15.19***} \\
 & Log OLS & 1\% & 38\% & \\
 & Equal-weighted & 79\% & 18\% & \\
 & Volume/dollar-weighted & 20\% & 76\% & \\
\midrule
\multirow{2}{*}{H4} & Dollar volume & 50\% & 80\% & \multirow{2}{*}{7.17***} \\
 & Share volume & 50\% & 20\% & \\
\midrule
\multirow{2}{*}{H5} & Level OLS & 69\% & 12\% & \multirow{2}{*}{10.01***} \\
 & Log OLS & 31\% & 88\% & \\
\midrule
\multirow{4}{*}{H6} & Price impact & 47\% & 74\% & \multirow{4}{*}{12.62***} \\
 & Amihud & 47\% & 26\% & \\
 & Level OLS & 49\% & 60\% & \\
 & Log OLS & 51\% & 40\% & \\
\midrule
\multirow{2}{*}{All} & Daily frequency & 93\% & 72\% & \multirow{2}{*}{$\chi^2$=14.31***} \\
 & Monthly frequency & 6\% & 28\% & \\
\bottomrule
\end{tabular}
\begin{minipage}{\textwidth}
\vspace{4pt}
\footnotesize\textit{Notes:} The test column reports the Anderson-Darling two-sample statistic comparing Sonnet and Opus effect size distributions for each hypothesis, and $\chi^2$ for the frequency row (agent-level attribute). *** denotes $p < 0.001$. All six hypotheses reject distributional equality. A consistent pattern emerges: Opus strongly prefers log OLS (H2: 86\%, H5: 88\%, H1: 64\%), while Sonnet prefers level OLS. For H3, Opus also favors volume/dollar-weighted aggregation (76\%) while Sonnet favors equal-weighted (79\%).
\end{minipage}
\end{table}

The most pronounced difference is in H1 (market efficiency): 87\% of Sonnet agents choose autocorrelation measures, while 100\% of Opus agents choose variance ratio measures. This is not a random draw; rather, it reflects stable preferences embedded in each model's training. The same pattern holds for functional form: Opus agents strongly prefer log OLS, while Sonnet agents prefer level OLS. For H3 (realized spread), Opus agents favor volume/dollar-weighted aggregation, while Sonnet agents favor equal-weighted. These differences suggest that different models have distinct empirical styles that shape their methodological choices and resulting estimates.

\section{Feedback Effects}
\label{sec:feedback}

\subsection{Peer Review Has Minimal Effect}

\begin{figure}[H]
\centering
\includegraphics[width=\textwidth]{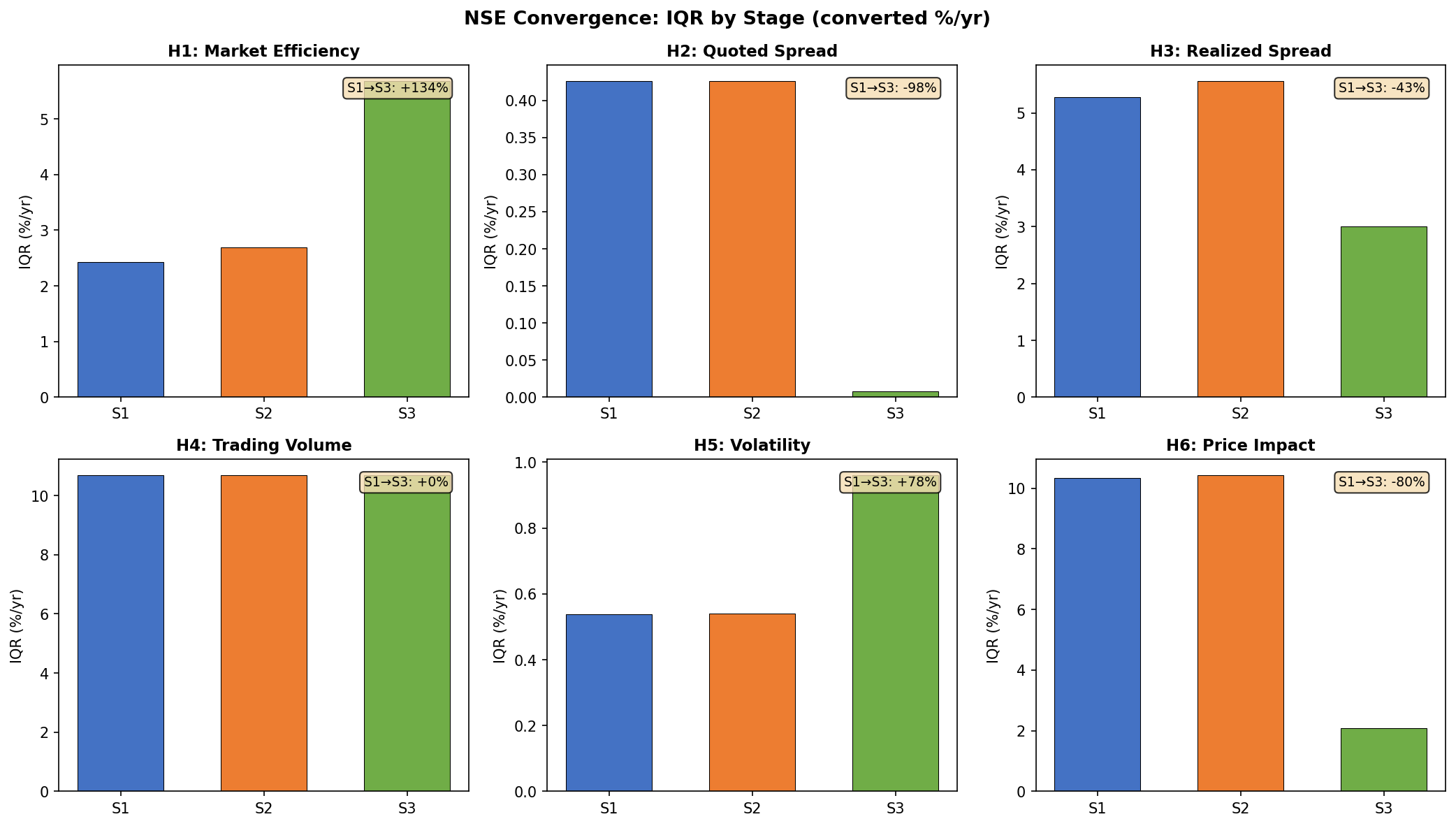}
\caption{IQR by stage for each hypothesis. Blue = S1, orange = S2, green = S3. The transition from S1 to S2 (peer review) produces minimal change. The transition from S2 to S3 (top papers) drives convergence for H2, H3, and H6, and divergence for H1 and H5.}
\label{fig:nse_convergence}
\end{figure}

Table~\ref{tab:convergence} and Figure~\ref{fig:nse_convergence} report the IQR at each stage.

\begin{table}[H]
\centering
\caption{IQR convergence across stages}
\label{tab:convergence}
\small
\begin{tabular}{lrrrrrrr}
\toprule
Hyp & S1 IQR & S2 IQR & S3 IQR & S1$\to$S2 & S2$\to$S3 & S1$\to$S3 \\
\midrule
H1 & 2.43 & 2.69 & 5.68 & $+$10.7\% & $+$111.1\% & $+$133.6\% \\
H2 & 0.43 & 0.43 & 0.01 & $+$0.2\% & $-$98.1\% & $-$98.1\% \\
H3 & 5.28 & 5.56 & 3.01 & $+$5.4\% & $-$46.0\% & $-$43.0\% \\
H4 & 10.69 & 10.69 & 10.69 & $+$0.0\% & $+$0.0\% & $+$0.0\% \\
H5 & 0.54 & 0.54 & 0.96 & $+$0.2\% & $+$78.1\% & $+$78.4\% \\
H6 & 10.34 & 10.44 & 2.09 & $+$1.0\% & $-$80.0\% & $-$79.8\% \\
\bottomrule
\end{tabular}
\end{table}

From S1 to S2 (peer review), the IQR is essentially unchanged for most hypotheses. On average, AI peer review does not reduce estimate dispersion.

Importantly, the absence of IQR reduction does not mean agents ignore the feedback. Between S1 and S2, 42\% of agents change their measure name, 29\% change their model specification, and 44\% change their effect size by more than 0.5\%/yr. The issue is that these changes are \textit{undirected}: agents move toward and away from the cross-agent median in roughly equal proportions (approximately 40\% each direction for H1, H4, and H6). Written critiques point out different issues for different agents, causing idiosyncratic changes that do not systematically reduce dispersion. Peer review generates movement but not convergence.

This contrasts with \citet{menkveld2024nonstandard}, where convergence is distributed roughly evenly across all four stages. The difference may reflect how human researchers and AI agents process feedback. Human researchers can evaluate a critique's merit, assess whether it applies to their specific approach, and make targeted adjustments. AI agents appear to treat feedback as a prompt for wholesale revision, sometimes switching measures entirely rather than refining their existing approach. The result is high individual-level volatility with no aggregate-level convergence.

The contrast with S3 (top papers) is instructive. Top papers cause \textit{correlated} movement: all agents see the same five exemplars and shift in the same direction, which mechanically reduces the IQR. Peer review causes \textit{uncorrelated} movement: each agent receives unique feedback and responds idiosyncratically, leaving the IQR unchanged.

\subsection{Top Papers Drive Dramatic Convergence}

From S1 to S3 (the combined effect of peer review and top-paper exposure), the IQR declines substantially for four of six hypotheses:
\begin{itemize}[nosep]
    \item H2: $-$98\% (IQR from 0.43 to 0.008)
    \item H6: $-$80\% (IQR from 10.34 to 2.09)
    \item H3: $-$43\% (IQR from 5.28 to 3.01; marginally significant, $p = 0.076$)
    \item H4:dollar\_volume: $-$97\% (stratified IQR from 0.25 to 0.007)
\end{itemize}
Nearly all of this convergence occurs in the S2-to-S3 transition (top papers), not the S1-to-S2 transition (peer review).

\begin{figure}[H]
\centering
\includegraphics[width=\textwidth]{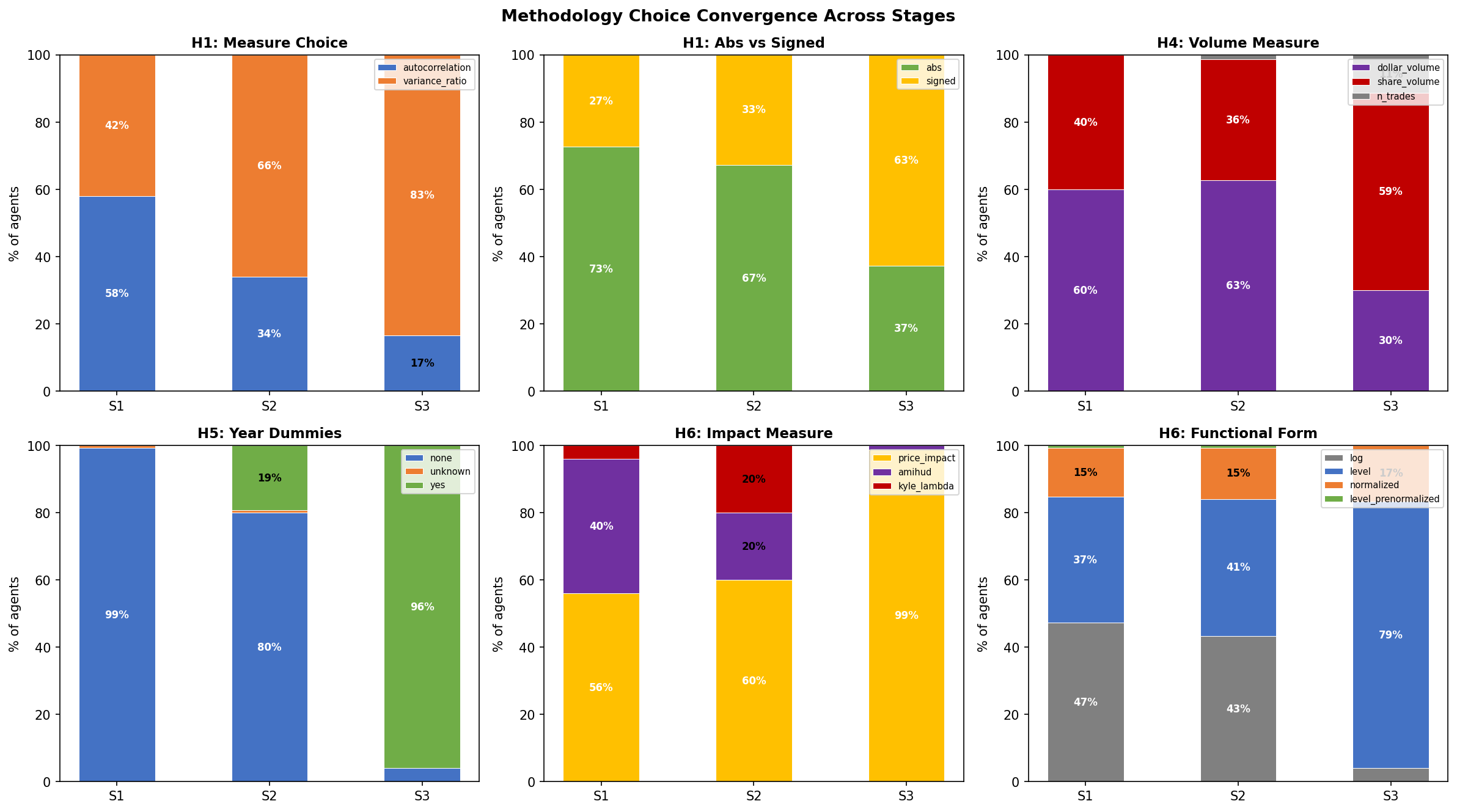}
\caption{Methodology choice convergence across stages. Each stacked bar shows the fraction of agents using each method. H5 year dummies: adoption rises from 1\% (S1) to 96\% (S3). H6 impact measure: price impact rises from 56\% to 99\%. H1 measure: autocorrelation drops from 58\% to 17\% as agents switch to variance ratio after seeing top papers.}
\label{fig:methodology_convergence}
\end{figure}

Figure~\ref{fig:methodology_convergence} shows the mechanism: agents shift their methodology to match the top papers. For H6, the fraction using trade-level price impact rises from 56\% (S1) to 99\% (S3), while Amihud illiquidity nearly disappears. For H5, year dummies for 2020/2022 are adopted by 96\% of agents in S3 (up from 1\% in S1).

\subsection{Stratified Convergence}

Table~\ref{tab:convergence_stratified} compares stratified convergence to the human benchmark from \citet{menkveld2024nonstandard}.

\begin{table}[H]
\centering
\caption{Stratified IQR convergence across stages}
\label{tab:convergence_stratified}
\small
\begin{tabular}{lrrrrrl}
\toprule
Family & $N_{\text{S1}}$ & $N_{\text{S3}}$ & S1 IQR & S3 IQR & $\Delta$IQR & $p$-value \\
\midrule
\multicolumn{7}{l}{\textit{Panel A: All agents (S1 and S3 populations may differ)}} \\
H1:autocorrelation & 87 & 25 & 2.74 & 0.13 & $-$95.4\% & \\
H1:variance\_ratio & 63 & 125 & 1.45 & 5.51 & $+$281.5\% & \\
H4:dollar\_volume & 90 & 45 & 0.25 & 0.007 & $-$97.0\% & \\
H4:share\_volume & 60 & 88 & 0.11 & 0.001 & $-$99.3\% & \\
\midrule
\multicolumn{7}{l}{\textit{Panel B: Stayers only (same agents in S1 and S3) and unstratified}} \\
H1:AC stayers & 25 & 25 & 7.16 & 0.13 & $-$98.3\% & 0.002$^{c}$ \\
H1:VR stayers & 63 & 63 & 1.45 & 2.50 & $+$73.2\% & 0.090$^{d}$ \\
H2 & 150 & 150 & 0.43 & 0.008 & $-$98.1\% & 0.005$^{c}$ \\
H3 & 150 & 150 & 5.28 & 3.01 & $-$43.0\% & 0.074$^{c}$ \\
H4:dollar stayers & 4 & 4 & 0.49 & 1.21 & $+$146\% & \\
H4:share stayers & 10 & 10 & 0.007 & 0.007 & $-$0.2\% & \\
H5 & 150 & 150 & 0.54 & 0.96 & $+$78.4\% & $<$0.001$^{d}$ \\
H6 & 150 & 150 & 10.34 & 2.09 & $-$79.8\% & $<$0.001$^{c}$ \\
\midrule
\textit{Menkveld (human, S1$\to$S4)} & \multicolumn{2}{c}{164} & & & $-$47.2\% & $<$0.005$^{c}$ \\
\bottomrule
\end{tabular}
\begin{minipage}{\textwidth}
\vspace{4pt}
\footnotesize\textit{Notes:} Panel A reports IQR using all agents assigned to each family at each stage; $N$ differs because agents switch families. Panel B restricts to ``stayers'' (agents using the same measure family in both S1 and S3) and unstratified hypotheses. $p$-values from bootstrap resampling (2,000 replicates): $^{c}$ tests $H_0$: IQR$_{\text{S3}} \geq$ IQR$_{\text{S1}}$ (convergence); $^{d}$ tests $H_0$: IQR$_{\text{S3}} \leq$ IQR$_{\text{S1}}$ (divergence). H4 stayer subgroups too small for reliable inference ($N = 4$ and $10$). The H1:VR stayer IQR increase ($+73\%$, $p = 0.090$) is driven by the signed-vs-abs sub-fork: signed agents find $+5.5$\%/yr while $|\text{VR}-1|$ agents find $-0.02$\%/yr. The H5 IQR increase ($+78\%$, $p < 0.001$) arises because year dummies interact differently with level OLS (median 0.21\%/yr) and log OLS (median 1.18\%/yr); within either specification, IQR $< 0.02$.
\end{minipage}
\end{table}

Within converging measure families, AI convergence from S1 to S3 ranges from $-$80\% to $-$99\%. For reference, \citet{menkveld2024nonstandard} report $-$47\% total convergence across four stages for human teams, though the comparison is not direct given the different hypotheses and datasets (see Section~\ref{sec:discussion}).

An important caveat is that convergence operates through two distinct mechanisms. The first is \textit{within-family estimation tightening}: agents that use the same measure in both S1 and S3 converge on nearly identical point estimates. The second is \textit{measure-family migration}: agents abandon their S1 measure choice and adopt the measure used in the top papers. For H1, 62 of 87 autocorrelation agents (71\%) switch to variance ratio by S3, leaving only 25 autocorrelation agents. For H4, measure-family transitions were bidirectional: 78 of 90 dollar-volume agents migrated to share volume, while 41 of 60 share-volume agents adopted dollar volume, and 17 agents adopted a third measure (number of trades) that no S1 agent used. The stage-specific sample sizes in Table~\ref{tab:convergence_stratified} reflect these compositional changes.

The remaining AI NSE after S3 is concentrated in two sources. First, unresolved measure-choice forks: H4 remains bimodal (dollar vs.\ share volume) because both top papers and the underlying research question admit both interpretations. Second, specification interaction effects: for H5, nearly all agents adopt year dummies from the top papers, but the dummies interact differently with the log vs.\ level specification, creating a bimodal split at approximately 0.21\%/yr (level) vs.\ approximately 1.18\%/yr (log).

\subsection{Divergence Cases}

Not all hypotheses converge. H1:variance\_ratio IQR \textit{increases} by 282\% from S1 to S3. The mechanism is instructive: two of the five top papers use signed variance ratio deviation $(\text{VR}(5) - 1)$ and find a positive trend. Agents that previously used autocorrelation switch to variance ratio but implement it inconsistently. Some use $|\text{VR} - 1|$ (absolute), finding near-zero trends; others use the signed version, finding approximately $+$5.5\%/yr. The top papers \textit{created} a new measure-choice fork rather than resolving an existing one.

\subsection{Measure Switching and Hypothesis Ambiguity}

A revealing pathology emerges for the abstract hypotheses (H1, H4, H6): agents switch measures in response to top papers even when the switch is not warranted. For H4, where the five top papers are split (2 share volume, 2 dollar volume, 1 number of trades), 91\% of agents switch their volume measure between S1 and S3. The transition matrix is striking: 78 of 90 dollar-volume agents switch to share volume, while simultaneously 41 of 60 share-volume agents switch to dollar volume, producing a near-random shuffle of family membership. Only 14 of 150 agents (9\%) retain their original measure.

This wholesale measure switching is \textit{economically irrational}. An agent that chose dollar volume in S1 because it measures capital flows has no analytical reason to switch to share volume in S3 merely because a top-rated paper used share volume. The switch reflects \textit{imitation without understanding}: agents copy the exemplar's measure choice without evaluating whether it is economically superior to their own.

For H6, the pattern is more rational. Four of five top papers use trade-level price impact; only one uses Amihud. In S3, 58 of 60 Amihud agents switch to price impact, while all 84 price-impact agents stay. This is directional convergence to the majority approach, not a random shuffle. But even here, the switch is debatable: the Amihud ratio and trade-level price impact measure different economic objects, and an agent's S1 choice of Amihud was not necessarily wrong.

The contrast between H4 (random shuffle) and H6 (directional convergence) reveals that agents do not evaluate whether measure switching is warranted. They simply imitate whichever top paper they attend to most. When the top papers are split on the measure choice (H4), the result is chaos; when they agree (H6), the result is convergence. In neither case does the agent reason about whether the alternative measure is economically more appropriate.

This finding has an important implication: for abstract hypotheses, exemplar exposure can \textit{degrade} the interpretability of the aggregate results even as it reduces the IQR within families. The S3 estimates are no more ``correct'' than the S1 estimates; they are merely more homogeneous within the dominant family.

\section{Multiverse Analysis}
\label{sec:multiverse}

\subsection{Fork Taxonomy}

Following the multiverse analysis framework of \citet{steegen2016increasing}, we extract approximately 30 methodological decision forks per agent by reading each agent's estimation code with a meta-agent. Table~\ref{tab:fork_taxonomy} classifies the key forks and their distributions.

\begin{table}[H]
\centering
\caption{Methodology fork taxonomy (Stage 1, 150 agents)}
\label{tab:fork_taxonomy}
\small
\begin{tabular}{llp{7cm}}
\toprule
Category & Fork & Distribution \\
\midrule
\multirow{4}{*}{\textbf{Measure}} & H1 measure family & autocorrelation 58\%, variance ratio 42\% \\
 & H1 abs vs.\ signed & absolute 73\%, signed 27\% \\
 & H4 measure family & dollar volume 60\%, share volume 40\% \\
 & H6 measure type & price impact 56\%, Amihud 40\%, Kyle $\lambda$ 4\% \\
\midrule
\multirow{4}{*}{\textbf{Estimation}} & Functional form & level OLS 56\%, log OLS 44\% \\
 & SE correction & HAC 59\%, NW 21\%, robust/HC3 11\%, none 9\% \\
 & Frequency & daily 86\%, monthly 13\%, annual 1\% \\
 & NW lag length & ranges from 2 to 252 (median 10) \\
\midrule
\multirow{4}{*}{\textbf{Data}} & Trading hours & RTH 9:30--16:00 (100\%) \\
 & Outlier treatment & none (85\%), winsorize (7\%), clip/trim (8\%) \\
 & Odd-lot inclusion & include 49\%, exclude 17\%, unknown 34\% \\
 & Trade subsampling & none (97\%), 5k--50k trades/day (3\%) \\
\bottomrule
\end{tabular}
\end{table}

The forks fall into three categories. \textit{Measure definition} (what quantity to compute) is the dominant source of AI NSE. \textit{Estimation specification} (functional form, SE correction, frequency) affects precision but rarely flips conclusions. \textit{Data handling} (trading hours, trade filtering, outlier treatment) is relatively homogeneous across agents. This hierarchy mirrors the structure identified by \citet{menkveld2024nonstandard} for human teams, where sampling frequency and model choice are the most consequential forks.

\subsection{Which Forks Matter Most?}

We quantify the contribution of each fork to estimate dispersion by regressing Stage 1 effect sizes on fork indicator variables (one fork at a time). Table~\ref{tab:variance_decomp} reports the univariate $R^2$ for key forks.

\begin{table}[H]
\centering
\caption{Variance decomposition: Which forks explain AI NSE?}
\label{tab:variance_decomp}
\small
\begin{tabular}{llrrl}
\toprule
Hypothesis & Fork & $R^2$ & $F$-stat & Sig. \\
\midrule
\multirow{4}{*}{H1: Efficiency} & abs\_or\_signed & 0.165 & 29.3 & *** \\
 & measure\_family & 0.165 & 29.1 & *** \\
 & llm (Sonnet/Opus) & 0.132 & 22.5 & *** \\
 & scale (log/level) & 0.085 & 4.5 & *** \\
\midrule
\multirow{3}{*}{H2: Quoted Spread} & scale (log/level) & 0.882 & 1110.8 & *** \\
 & llm (Sonnet/Opus) & 0.446 & 119.0 & *** \\
 & se\_correction & 0.095 & 5.1 & *** \\
\midrule
\multirow{3}{*}{H3: Realized Spread} & trade\_weighting & 0.145 & 8.2 & *** \\
 & scale (log/level) & 0.027 & 4.1 & ** \\
 & llm (Sonnet/Opus) & 0.021 & 3.1 & * \\
\midrule
\multirow{3}{*}{H4: Volume} & measure\_family & 0.954 & 3069.5 & *** \\
 & llm (Sonnet/Opus) & 0.085 & 13.8 & *** \\
 & scale (log/level) & 0.054 & 8.5 & *** \\
\midrule
\multirow{2}{*}{H5: Volatility} & scale (log/level) & 0.205 & 38.2 & *** \\
 & se\_correction & 0.023 & 1.1 & \\
\midrule
\multirow{4}{*}{H6: Price Impact} & h6\_measure\_type & 0.623 & 121.6 & *** \\
 & trade\_weighting & 0.478 & 33.2 & *** \\
 & scale (log/level) & 0.356 & 81.8 & *** \\
 & llm (Sonnet/Opus) & 0.065 & 10.3 & *** \\
\bottomrule
\end{tabular}
\begin{minipage}{\textwidth}
\vspace{4pt}
\footnotesize\textit{Notes:} Each row reports the $R^2$ from a univariate OLS regression of Stage 1 effect sizes (\%/yr) on indicator variables for the fork. ``All combined'' includes all forks jointly. ***, **, * denote significance at 1\%, 5\%, 10\%.
\end{minipage}
\end{table}

The results are notable. For H4 (volume), the measure\_family fork alone explains 95.4\% of all cross-agent variation. For H6 (price impact), h6\_measure\_type explains 62.3\% and trade\_weighting explains 47.8\%. For H2 (quoted spread), the scale fork (log vs.\ level) explains 88.2\%, indicating that almost all variation is in the \textit{units} of the estimate rather than the economic magnitude. For H5, the scale fork explains 20.5\%. For H1, the abs\_or\_signed fork explains 16.5\%, and the LLM backbone (Sonnet vs.\ Opus) explains 13.2\%.

\section{Discussion}
\label{sec:discussion}

\subsection{Structure of AI NSE in Context}

Table~\ref{tab:ai_vs_human} contextualizes our findings alongside the human NSE results of \citet{menkveld2024nonstandard}, noting that the comparison is suggestive given the different hypotheses and datasets.

\begin{table}[H]
\centering
\caption{AI NSE in context: comparison with \citet{menkveld2024nonstandard}}
\label{tab:ai_vs_human}
\small
\begin{tabular}{lp{5cm}p{5cm}}
\toprule
Property & Human \citep{menkveld2024nonstandard} & AI (this paper) \\
\midrule
Paradigm diversity & High (linear trend 35\%, relative changes 58\%, log-diff 6\%) & Low (level regression 56\%, log regression 44\%; no relative changes) \\
Measure diversity & Moderate & High (model-dependent) \\
Jensen's inequality & Major source of outliers & Absent \\
Feedback response & Gradual, each stage contributes & Binary: exemplars work, critiques do not \\
PE quality $\to$ NSE & Higher quality $\to$ $-$33\% IQR & Mixed (Appendix~\ref{app:pe_quality}): significant for H4, H6; not for H1, H2, H5 \\
Convergence S1$\to$S3/S4 & $-$47\% IQR (4 stages) & $-$80\% to $-$99\% stratified (3 stages) \\
\bottomrule
\end{tabular}
\end{table}

The table reveals that AI NSE is concentrated in measure-choice forks with near-zero within-family variation, whereas \citet{menkveld2024nonstandard} find that human NSE is additionally driven by estimation-paradigm diversity. However, we caution against direct comparisons: the hypotheses, datasets, and instruments differ, and we do not have access to the human-team data.

An important distinction emerges between two types of cross-agent variation. The first is the variation in estimates arising from different but defensible analytical choices applied to the same well-defined construct (e.g., using robust standard errors, or daily vs.\ monthly frequency). The second is \textit{hypothesis abstraction}: variation arising because the hypothesis itself admits multiple valid interpretations (e.g., ``market efficiency'' can be measured via autocorrelation or variance ratio; ``trading volume'' can mean dollar volume or share volume). For the three abstract hypotheses (H1, H4, H6), much of the cross-agent variation reflects hypothesis abstraction rather than analytical disagreement. Within each interpretation, agents converge tightly. This distinction is important because hypothesis abstraction is a property of the research question, not of the researcher or agent, and cannot be reduced through feedback.

Several caveats apply to the comparison with \citet{menkveld2024nonstandard}. First, the six hypotheses are not the same: only H1 (market efficiency) is shared across both studies. Menkveld's H2--H6 concern client-dealer dynamics in EURO STOXX 50 futures (realized spread, client volume share, client market orders, gross trading revenue), while our H2--H6 concern broader market quality measures for SPY (quoted spread, realized spread, volume, volatility, price impact). The convergence comparison is therefore structural (same feedback protocol, similar number of hypotheses) rather than hypothesis-by-hypothesis. Second, their teams analyze a single-venue futures market (Eurex) while our agents analyze a fragmented equity ETF (SPY across 16+ venues), creating different analytical decision spaces. Third, the \citet{menkveld2024nonstandard} convergence figure of $-47\%$ spans four stages including a Bayesian update stage that we omit.

For H1, the one shared hypothesis, both human and AI researchers face the same measure-choice fork: autocorrelation vs.\ variance ratio. Human teams split 37\%/63\% (autocorrelation/VR); our AI agents split 58\%/42\%. The key difference is that human teams additionally diverge on the estimation paradigm: 58\% use relative changes (period-over-period ratios), a specification that AI agents never select.

\subsection{AI NSE as a Lower Bound on Human NSE}

An important implication of our findings is that AI NSE potentially provides a \textit{lower bound} on the nonstandard errors that would arise among human researchers facing the same task.

The argument is as follows. AI agents share a common training corpus, a common architecture, and common instruction-following behavior. The only sources of variation are sampling stochasticity and model-family differences. Human researchers, by contrast, bring idiosyncratic graduate training, institutional norms, disciplinary preferences, career incentives, and varying levels of domain expertise. Every fork that AI agents disagree on (e.g., dollar vs.\ share volume, autocorrelation vs.\ variance ratio) is a fork that human researchers would also disagree on, because the disagreement reflects genuine ambiguity in the research question rather than any deficiency in the analyst. Human researchers may additionally disagree on forks where AI agents show limited diversity, such as the relative-change paradigm, which \citet{menkveld2024nonstandard} identify as a major source of dispersion. We cannot verify this conjecture directly because we do not have human-team data on our hypotheses, but the logic suggests that AI NSE provides a plausible lower bound.

This framing reinterprets AI NSE not as a deficiency but as a diagnostic. The measure-choice forks that AI agents cannot resolve (H4: dollar vs.\ share volume; H6: price impact vs.\ Amihud) reveal irreducible ambiguity in the research question itself. No amount of methodological training or feedback can resolve a disagreement that stems from the hypothesis being underspecified. The specification-choice forks (H2: log vs.\ level; H5: log vs.\ level with dummies) reveal sensitivity to analytical details that would persist in any population of researchers. Paradigm-choice variation (e.g., relative changes) may be specific to human researchers and could in principle be reduced through standardization.

This perspective suggests a practical use for AI agents in empirical research: deploying a multiverse of AI agents as a \textit{pre-registration diagnostic} to estimate the minimum NSE for a given research design before committing human effort. If AI agents already disagree substantially, the research question likely needs further specification. If they agree, the researcher can proceed with greater confidence that human NSE will be manageable.

\subsection{Implications for Automated Policy Evaluation}

Projects such as APE \citep{ape2026} deploy AI agents to produce end-to-end economics research. Our findings suggest that the credibility of AI-generated empirical results depends critically on which measure-choice forks the agent happens to take, and different model families have different stable preferences. A single AI-generated estimate should not be treated as ground truth.

We recommend \textit{multiverse analysis} \citep{steegen2016increasing} as a standard complement to AI-generated estimates: running the same analysis with multiple measure definitions, model specifications, and agent configurations to reveal the full distribution of plausible results. This recommendation aligns with recent calls for specification transparency in empirical research \citep{simmons2011false,brodeur2020methods}.

\subsection{Implications for AI-Assisted Research}

For researchers using AI as coding assistants, our findings highlight that the tool's defaults encode methodological preferences. A researcher using Sonnet would likely see autocorrelation-based efficiency measures; the same researcher using Opus would see variance ratio measures. Reporting the LLM model version is necessary but insufficient for reproducibility, as the same model at different temperature draws produces different fork choices. This concern echoes the broader literature on how researcher degrees of freedom influence empirical results \citep{simmons2011false}, with the added complication that the ``researcher'' is now an LLM whose degrees of freedom are opaque even to its developers.

\subsection{AI NSE as Learned Uncertainty, Not Error}

The term ``nonstandard errors'' frames analytical variation as a problem to be eliminated. We argue for a more nuanced interpretation: much of the variation we observe reflects \textit{inherent uncertainty in social science research design} that AI agents have learned from the research literature itself.

During pre-training, large language models process millions of academic papers spanning decades of methodological debate. An LLM that has read papers using autocorrelation \textit{and} papers using variance ratios to measure market efficiency has, in effect, learned that the field does not agree on a single operationalization. When different agent runs select different measures, they are sampling from the empirical distribution of defensible analytical choices as represented in the training corpus. In this view, AI NSE is not a deficiency of the model but a faithful reflection of the methodological uncertainty that the social science literature itself contains.

This interpretation has an important implication for model development. As LLM providers invest in post-training alignment (RLHF, constitutional AI, instruction tuning), there may be pressure to reduce output variability by steering models toward ``canonical'' approaches. Our findings suggest this would be counterproductive for research applications. If a model always chose dollar volume over share volume, or always used autocorrelation over variance ratio, it would appear more consistent but would mask genuine uncertainty about which operationalization is appropriate. The resulting estimates would be precise but potentially misleading, since the apparent consensus would reflect model alignment rather than scientific agreement.

We therefore recommend that AI NSE be \textit{preserved rather than eliminated} in models intended for research use. The variation across runs is informative: it reveals which aspects of a research question are well-specified (H2: all agents agree on quoted spread) and which are underspecified (H4: agents split on dollar vs.\ share volume). Deployed as a multiverse tool, AI agents can provide a rapid, low-cost assessment of the inherent uncertainty in a research design before human researchers commit effort to a single analytical path.

\subsection{Limitations}

Several limitations warrant discussion.

\textit{Trade direction classification.} The realized spread (H3) and price impact (H6) measures require classifying each trade as buyer- or seller-initiated, typically via the Lee-Ready algorithm \citep{lee1991inferring}. For SPY, where trades occur at sub-millisecond frequency and the spread is often 1 cent, tick-test classification is known to perform poorly. No agent validates classification accuracy, and the conversion pipeline does not adjust for potential misclassification bias.

\textit{Effective spread decomposition.} In the standard microstructure framework, the effective spread decomposes approximately into realized spread plus price impact: $\text{H2} \approx \text{H3} + \text{H6}$ \citep{bessembinder1997comparison}. No agent verifies this internal consistency condition, and we do not test whether the H2, H3, and H6 estimates satisfy it.

\textit{Top paper selection and evaluator bias.} The five exemplar papers shown in Stage 3 are selected by AI peer evaluation scores. If AI evaluators systematically prefer certain measure choices (e.g., Opus evaluators preferring variance ratio, which Opus agents exclusively use), the top paper selection may propagate evaluator preferences rather than reflect genuine quality. The S3 convergence would then partly reflect evaluator bias rather than methodological learning. We do not formally test for this confound.

\textit{External validity.} We examine a single, extremely liquid asset (SPY) in a single domain (market microstructure). SPY's microstructure is well-documented in the finance literature and likely well-represented in LLM training data, which may narrow the space of defensible approaches relative to less-studied settings. Whether AI NSE patterns generalize to other empirical domains remains an open question.


\section{Conclusion}
\label{sec:conclusion}

This paper documents that state-of-the-art AI coding agents exhibit sizable nonstandard errors.
When given the same research question and dataset, different agent runs produce widely varying estimates of key market quality measures for SPY. The variation is not random noise but structured: it arises from specific methodological forks (e.g., autocorrelation vs.\ variance ratio for market efficiency) that are stable across runs and model families.
A common practice, AI peer review (S2), is surprisingly ineffective at reducing the NSE in AI agents as it fails to resolve the underlying methodological forks and may encourage agents to explore different methodological paths. 
When exposed to exemplar papers, AI agents converge through two mechanisms: within-family estimation tightening (IQR reductions of 80--99\% in converging families) and cross-family migration (e.g., 71\% of autocorrelation agents switching to variance ratio). However, exemplar exposure can also \textit{increase} dispersion when top papers introduce new methodological options that agents adopt inconsistently.

These findings carry several implications for the growing use of AI in automated research. First, a single AI estimate should not be trusted as definitive, since the ``which measure'' fork alone can flip conclusions. Second, AI peer review is ineffective at reducing NSE, while exemplar-based calibration is far more powerful but can backfire. Third, different model families encode different methodological ``styles'' that persist across runs. Fourth, multiverse analysis should be standard practice when deploying AI agents for empirical research.

Our study has limitations. We examine a single asset (SPY) in a single domain (market microstructure).  While we are hopeful, whether AI NSE patterns generalize to other empirical settings remains an open question. Future work could extend this framework to other domains, examine additional model families, and develop automated multiverse tools that systematically explore the space of defensible analytical choices.

\newpage


\paragraph{Data and code availability.} All agent outputs (research reports, results CSVs), the converted results, and the analysis scripts used in this paper are available at \url{https://github.com/ruijiang81/AI_NSE/}. The underlying NYSE TAQ data can be accessed through \href{https://wrds-www.wharton.upenn.edu/}{Wharton Research Data Services}, but the agent outputs and analysis code are sufficient to reproduce all tables and figures.

\bibliographystyle{apalike}
\bibliography{references}

@article{amihud2002illiquidity,
  author  = {Amihud, Yakov},
  title   = {Illiquidity and Stock Returns: Cross-Section and Time-Series Effects},
  journal = {Journal of Financial Markets},
  volume  = {5},
  number  = {1},
  pages   = {31--56},
  year    = {2002}
}

@article{bessembinder1997comparison,
  title={A comparison of trade execution costs for NYSE and NASDAQ-listed stocks},
  author={Bessembinder, Hendrik and Kaufman, Herbert M},
  journal={Journal of Financial and Quantitative Analysis},
  volume={32},
  number={3},
  pages={287--310},
  year={1997},
  publisher={Cambridge University Press}
}

@misc{ape2026,
  author       = {{Social Catalyst Lab}},
  title        = {Project {APE}: Autonomous Policy Evaluation},
  year         = {2026},
  howpublished = {University of Zurich, Department of Economics},
  url          = {https://ape.socialcatalystlab.org/}
}

@article{argyle2023out,
  author  = {Argyle, Lisa P. and Busby, Ethan C. and Fulda, Nancy and Gubler, Joshua R. and Rytting, Christopher and Wingate, David},
  title   = {Out of One, Many: Using Language Models to Simulate Human Samples},
  journal = {Political Analysis},
  volume  = {31},
  number  = {3},
  pages   = {337--351},
  year    = {2023}
}

@article{botvinik2020variability,
  author  = {Botvinik-Nezer, Rotem and Holzmeister, Felix and Camerer, Colin F. and Dreber, Anna and Huber, Juergen and Johannesson, Magnus and others},
  title   = {Variability in the Analysis of a Single Neuroimaging Dataset by Many Teams},
  journal = {Nature},
  volume  = {582},
  number  = {7810},
  pages   = {84--88},
  year    = {2020}
}

@article{breznau2022observing,
  author  = {Breznau, Nate and Rinke, Eike Mark and Wuttke, Alexander and others},
  title   = {Observing Many Researchers Using the Same Data and Hypothesis Reveals a Hidden Universe of Uncertainty},
  journal = {Proceedings of the National Academy of Sciences},
  volume  = {119},
  number  = {44},
  pages   = {e2203150119},
  year    = {2022}
}

@article{brodeur2020methods,
  author  = {Brodeur, Abel and Cook, Nikolai and Heyes, Anthony},
  title   = {Methods Matter: p-Hacking and Publication Bias in Causal Analysis in Economics},
  journal = {American Economic Review},
  volume  = {110},
  number  = {11},
  pages   = {3634--3660},
  year    = {2020}
}

@article{harvey2016and,
  author  = {Harvey, Campbell R. and Liu, Yan and Zhu, Heqing},
  title   = {\ldots and the Cross-Section of Expected Returns},
  journal = {The Review of Financial Studies},
  volume  = {29},
  number  = {1},
  pages   = {5--68},
  year    = {2016}
}

@techreport{horton2023homo,
  author      = {Horton, John J. and Filippas, Apostolos and Manning, Benjamin},
  title       = {Large Language Models as Simulated Economic Agents: What Can We Learn from Homo Silicus?},
  institution = {National Bureau of Economic Research},
  type        = {Working Paper},
  number      = {31122},
  year        = {2023}
}

@unpublished{huang2026ai,
  author = {Huang, Wenqian and Menkveld, Albert J. and Yu, Shihao},
  title  = {{AI} ``Errors''},
  note   = {Working paper, Bank for International Settlements, Vrije Universiteit Amsterdam, and Singapore Management University},
  year   = {2026}
}

@article{lee1991inferring,
  author  = {Lee, Charles M. C. and Ready, Mark J.},
  title   = {Inferring Trade Direction from Intraday Data},
  journal = {The Journal of Finance},
  volume  = {46},
  number  = {2},
  pages   = {733--746},
  year    = {1991}
}

@article{lo1988stock,
  author  = {Lo, Andrew W. and MacKinlay, A. Craig},
  title   = {Stock Market Prices Do Not Follow Random Walks: Evidence from a Simple Specification Test},
  journal = {The Review of Financial Studies},
  volume  = {1},
  number  = {1},
  pages   = {41--66},
  year    = {1988}
}

@article{lu2024aiscientist,
  author  = {Lu, Chris and Lu, Cong and Lange, Robert Tjarko and Foerster, Jakob and Clune, Jeff and Ha, David},
  title   = {The {AI} Scientist: Towards Fully Automated Open-Ended Scientific Discovery},
  journal = {arXiv preprint arXiv:2408.06292},
  year    = {2024}
}

@article{menkveld2024nonstandard,
  author  = {Menkveld, Albert J. and Dreber, Anna and Holzmeister, Felix and Huber, Juergen and Johannesson, Magnus and Kirchler, Michael and others},
  title   = {Nonstandard Errors},
  journal = {The Journal of Finance},
  volume  = {79},
  number  = {3},
  pages   = {2339--2390},
  year    = {2024}
}

@inproceedings{perez2023discovering,
  author    = {Perez, Ethan and Ringer, Sam and Luko{\v{s}}i{\={u}}t{\.{e}}, Kamil{\.{e}} and Nguyen, Karina and Chen, Edwin and Heiner, Scott and others},
  title     = {Discovering Language Model Behaviors with Model-Written Evaluations},
  booktitle = {Findings of ACL 2023},
  pages     = {13387--13434},
  year      = {2023}
}

@inproceedings{sharma2024towards,
  author    = {Sharma, Mrinank and Tong, Meg and Korbak, Tomasz and Duvenaud, David and Askell, Amanda and Bowman, Samuel R. and others},
  title     = {Towards Understanding Sycophancy in Language Models},
  booktitle = {Proceedings of ICLR 2024},
  year      = {2024}
}

@article{silberzahn2018many,
  author  = {Silberzahn, Raphael and Uhlmann, Eric L. and Martin, Daniel P. and Anselmi, Pasquale and Aust, Frederik and Awtrey, Eli and others},
  title   = {Many Analysts, One Data Set: Making Transparent How Variations in Analytic Choices Affect Results},
  journal = {Advances in Methods and Practices in Psychological Science},
  volume  = {1},
  number  = {3},
  pages   = {337--356},
  year    = {2018}
}

@article{simmons2011false,
  author  = {Simmons, Joseph P. and Nelson, Leif D. and Simonsohn, Uri},
  title   = {False-Positive Psychology: Undisclosed Flexibility in Data Collection and Analysis Allows Presenting Anything as Significant},
  journal = {Psychological Science},
  volume  = {22},
  number  = {11},
  pages   = {1359--1366},
  year    = {2011}
}

@article{steegen2016increasing,
  author  = {Steegen, Sara and Tuerlinckx, Francis and Gelman, Andrew and Vanpaemel, Wolf},
  title   = {Increasing Transparency Through a Multiverse Analysis},
  journal = {Perspectives on Psychological Science},
  volume  = {11},
  number  = {5},
  pages   = {702--712},
  year    = {2016}
}

@inproceedings{zhang2025invisible,
  title={The invisible hand: Unveiling provider bias in large language models for code generation},
  author={Zhang, Xiaoyu and Zhai, Juan and Ma, Shiqing and Bao, Qingshuang and Jiang, Weipeng and Wang, Qian and Shen, Chao and Liu, Yang},
  booktitle={Proceedings of the 63rd Annual Meeting of the Association for Computational Linguistics (Volume 1: Long Papers)},
  pages={21376--21403},
  year={2025}
}

\newpage
\begin{appendices}
\section{Agent Instructions}
\label{app:instructions}

\subsection{Cost, Running Time, and Turns}

Table~\ref{tab:cost} reports the API cost, wall-clock running time, and number of conversational turns per agent-stage. The total cost of the experiment is \$1,558 across 450 agent-stage runs.

\begin{table}[H]
\centering
\caption{Cost, running time, and turns per agent-stage}
\label{tab:cost}
\footnotesize
\begin{tabular}{llrrrr}
\toprule
& & Mean & Median & Min & Max \\
\midrule
\multicolumn{6}{l}{\textit{API cost (\$ per agent-stage)}} \\
& Sonnet S1 & 3.54 & 2.85 & 1.30 & 14.03 \\
& Sonnet S2 & 2.13 & 1.58 & 0.69 & 13.63 \\
& Sonnet S3 & 3.12 & 2.56 & 1.27 & 10.52 \\
& Opus S1 & 5.59 & 4.19 & 1.83 & 18.70 \\
& Opus S2 & 3.21 & 2.57 & 1.31 & 19.05 \\
& Opus S3 & 4.75 & 3.56 & 1.44 & 20.03 \\
\midrule
\multicolumn{6}{l}{\textit{Wall-clock time (minutes per agent-stage)}} \\
& S1 & 60 & 53 & 21 & 119 \\
& S2 & 31 & 27 & 6 & 98 \\
& S3 & 31 & 25 & 6 & 110 \\
\midrule
\multicolumn{6}{l}{\textit{Number of turns per agent-stage}} \\
& S1 & 82 & 66 & 37 & 387 \\
& S2 & 54 & 44 & 26 & 388 \\
& S3 & 58 & 51 & 30 & 278 \\
\midrule
\multicolumn{6}{l}{\textit{Totals}} \\
& Total API cost & \multicolumn{4}{l}{\$1,558 (S1: \$634, S2: \$374, S3: \$550)} \\
& Total wall time & \multicolumn{4}{l}{$\sim$122 min/agent $\times$ 150 agents $\div$ 12 parallel = $\sim$25 hours} \\
\bottomrule
\end{tabular}
\begin{minipage}{\textwidth}
\vspace{4pt}
\footnotesize\textit{Notes:} API cost reflects Anthropic's pricing for Sonnet 4.6 and Opus 4.6 as of March 2026. Opus is approximately 1.6$\times$ more expensive than Sonnet per stage. Wall-clock time includes data processing (reading 66 GB of TAQ parquet files), code writing, and report generation. The \$20 budget cap per agent-stage was rarely binding (max cost \$20.03 for one Opus S3 run).
\end{minipage}
\end{table}

\subsection{System Setup and Pipeline}

\paragraph{Infrastructure.} All experiments run on a shared HPC cluster managed by SLURM. Each agent is submitted as an array task, with up to 12 agents running in parallel on a single compute node (40 GB RAM, 6 CPUs per agent). No GPU is used; all computation is CPU-based.

\paragraph{Container isolation.} Each agent runs inside a Singularity container built from a custom Docker image containing Python 3.12, the Claude Code CLI, and standard scientific Python packages. The container enforces filesystem isolation: each agent has a private read-write workspace, read-only access to the shared TAQ data directory, and no access to other agents' workspaces. This ensures that variation across agents arises solely from the stochasticity of the language model, not from shared state or communication.

\paragraph{Agent invocation.} Within the container, the Claude Code CLI runs in fully autonomous mode with a \$20 budget cap per stage. The agent has unrestricted permissions to read files, write code, execute programs, and manage processes within its workspace. All tool calls, assistant messages, and cost information are captured in structured JSONL log files for post-hoc analysis.

\paragraph{Three-stage pipeline.} The pipeline proceeds in five steps:
\begin{enumerate}[nosep]
\item \textit{Stage 1}: Each agent receives an environment specification, research instructions (Appendix~\ref{app:instructions}.3), and read-only access to the TAQ data. It independently produces a structured results file, a research report (2,000--4,000 words), analysis code, and figures.
\item \textit{Peer evaluation}: Two AI evaluators (one Sonnet, one Opus) read each agent's Stage 1 report and produce per-hypothesis ratings (0--10) and written feedback. Evaluators run in separate containers with read-only access to the agent's output.
\item \textit{Stage 2}: Each agent receives its two anonymized peer evaluations and revises its analysis.
\item \textit{Top paper selection}: The five agents with the highest average Stage 2 peer evaluation scores are selected. Their anonymized reports and results are distributed to all agents.
\item \textit{Stage 3}: Each agent receives the five top papers and may revise its analysis one final time.
\end{enumerate}

\paragraph{Effect size conversion pipeline.} Because agents report effect sizes in heterogeneous units (level OLS in raw measure units vs.\ log OLS in approximate \%/yr), we develop a per-agent conversion pipeline. For each of the 150 research agents, a separate Opus 4.6 conversion agent is launched as a SLURM array task. The conversion agent receives a structured prompt specifying the task (classify models, convert units, extract methodology forks) and read-only access to the research agent's workspace (code, results, and processed data). It cannot modify the research agent's files.

The conversion agent performs three tasks. First, it reads the research agent's results file and classifies each hypothesis's model as log, level, or normalized, using both the model name and a magnitude sanity check (e.g., a level-model H2 effect of $-0.000237$ is orders of magnitude smaller than a log-model effect of $-6.63$). Second, for level-model hypotheses, it reads the agent's processed daily measures, identifies the matching column via keyword matching, and computes the denominator for conversion to \%/yr. For measures that oscillate around zero (e.g., AR(1) autocorrelation with $\sim$35\% negative daily values, or realized spread), it uses the mean of absolute values as the denominator to avoid division by a near-zero residual mean. It also reads the estimation code to detect pre-scaling (e.g., agents that multiply the regression slope by 100 before reporting). Third, it reads the agent's Python source code to extract approximately 30 methodology fork annotations per hypothesis (measure family, functional form, SE correction, Newey-West lag count, trade weighting, reversal horizon, outlier treatment, trading hours, odd-lot inclusion, etc.).

Each conversion agent processes all three stages for its assigned research agent in a single session, producing one CSV per stage with 49 columns. The total cost of the conversion pipeline is approximately \$240 (150 agents $\times$ \$1.60 per agent). All conversion outputs are saved to a shared output directory and serve as the input to the analysis in this paper.

\subsection{Agent Instructions}

The following instructions were provided verbatim to all 150 agents. Each agent also received a \texttt{CLAUDE.md} file specifying the computing environment, data paths, output format, and deliverable structure (not reproduced here; available in the repository).

\subsubsection{Assignment}

You are expected to write a research report (2,000--4,000 words). For each of the six hypotheses listed below, you should:
\begin{enumerate}[nosep]
\item Propose a statistical measure, briefly motivate it, and present the formula to calculate it.
\item For this measure, estimate the average per-year change in percentage points, based on the full sample (or at least the longest possible period, because some series may not be meaningful at the beginning of the sample). Test it against the null of no change.
\item Report this estimate along with its standard error and the direction of the effect.
\item Briefly discuss your result.
\end{enumerate}

For example, an appropriate outcome statement for testing hypothesis X which states that Y has not changed is:

\begin{quote}
``We propose measure Z to test hypothesis X because [...]. It is calculated as Z = f(DATA). Implementing it leads to the following result: We reject the null of no change. We find that Y declined as our measure Z declined by 1.251\% on average per year where the standard error of this change is 0.421\% and the resulting t-statistic is 2.971. This result shows [...]''
\end{quote}

All results are in \textbf{percentage points per year} (e.g., a 1.234\% annual decline is reported as $-1.234$, not $-0.01234$).

\subsubsection{Hypotheses}

\textbf{H1: Market Efficiency.}
Assuming that informationally-efficient prices follow a random walk, did market efficiency change over time?

\textit{Null hypothesis 1:} Market efficiency has not changed over the sample period (2015--2024).

\medskip\noindent
\textbf{H2: Quoted Bid-Ask Spread.}
The quoted bid-ask spread is the difference between the best ask and the best bid price. It is a standard measure of trading cost. Did the quoted bid-ask spread change over time?

\textit{Null hypothesis 2:} The quoted bid-ask spread has not changed over the sample period (2015--2024).

\medskip\noindent
\textbf{H3: Realized Bid-Ask Spread.}
The realized spread could be thought of as the gross-profit component of the spread as earned by the liquidity provider (limit-order submitter). It compares the trade price to the midpoint some time after the trade, thereby capturing the portion of the spread that is not lost to adverse selection. Did the realized bid-ask spread change over time?

\textit{Null hypothesis 3:} The realized bid-ask spread has not changed over the sample period (2015--2024).

\medskip\noindent
\textbf{H4: Trading Volume.}
Did daily trading volume change over time?

\textit{Null hypothesis 4:} Daily trading volume has not changed over the sample period (2015--2024).

\medskip\noindent
\textbf{H5: Intraday Volatility.}
Intraday price volatility captures the magnitude of price fluctuations within a trading day. Did intraday volatility change over time?

\textit{Null hypothesis 5:} Intraday volatility has not changed over the sample period (2015--2024).

\medskip\noindent
\textbf{H6: Price Impact of Trades.}
Price impact measures how much prices move in response to trading activity. It captures the information content of trades and the depth of the market. Did the price impact of trades change over time?

\textit{Null hypothesis 6:} The price impact of trades has not changed over the sample period (2015--2024).

\section{Does AI Peer Evaluation Quality Predict NSE?}
\label{app:pe_quality}

\citet{menkveld2024nonstandard} find that higher peer evaluation (PE) ratings are associated with lower NSE: a one-standard-deviation increase in PE rating reduces the IQR by 33\%. We test whether a similar relationship holds for AI agents.

Each agent's Stage 1 report is evaluated by two AI peer evaluators (one Sonnet, one Opus) on a 0--10 scale per hypothesis. We compute the average rating per agent per hypothesis and correlate it with the absolute deviation of the agent's effect size from the cross-agent median. Table~\ref{tab:pe_nse} reports Spearman rank correlations and IQR by rating tercile.

\begin{table}[H]
\centering
\caption{PE rating vs.\ NSE: Spearman correlation and IQR by rating tercile (Stage 1)}
\label{tab:pe_nse}
\small
\begin{tabular}{lrrrrr}
\toprule
Hyp & Spearman $r$ & $p$-value & Low-rated IQR & Mid-rated IQR & High-rated IQR \\
\midrule
H1 & $-$0.040 & 0.628 & 2.45 & 2.31 & 1.09 \\
H2 & $+$0.125 & 0.127 & 0.06 & 0.53 & 0.38 \\
H3 & $-$0.189 & 0.021** & 5.83 & 3.81 & 3.18 \\
H4 & $-$0.325 & $<$0.001*** & 10.60 & 10.70 & 0.82 \\
H5 & $-$0.099 & 0.229 & 0.54 & 0.54 & 0.48 \\
H6 & $-$0.601 & $<$0.001*** & 5.77 & 11.88 & 2.44 \\
\bottomrule
\end{tabular}
\begin{minipage}{\textwidth}
\vspace{4pt}
\footnotesize\textit{Notes:} Spearman $r$ is the rank correlation between per-hypothesis PE rating (average of two evaluators) and $|\text{effect\_size} - \text{median}|$. Negative $r$ means higher-rated agents are closer to the median. IQR computed within rating terciles. **, *** denote $p < 0.05$, $p < 0.01$.
\end{minipage}
\end{table}

The relationship between PE quality and NSE is significant only for hypotheses with measure-choice ambiguity: H4 ($r = -0.325$, $p < 0.001$) and H6 ($r = -0.601$, $p < 0.001$). For H4, high-rated agents have an IQR of 0.82 (vs.\ 10.60 for low-rated), because AI evaluators give higher scores to agents using the more common measure family (dollar volume), which mechanically places them closer to the median. For H6, the pattern is similar: high-rated agents disproportionately use trade-level price impact rather than Amihud.

For well-specified hypotheses (H2, H5), PE ratings do not predict deviation because there is little deviation to predict (IQR $\leq 0.54$ regardless of rating). For H1, the relationship is weak and insignificant.

This contrasts with \citet{menkveld2024nonstandard}, where PE quality predicts NSE across all six hypotheses. The difference likely reflects the structural distinction between human and AI NSE and reviews: human NSE arises from a continuous distribution of analytical competence, which PE ratings capture; AI NSE arises from discrete measure-choice forks, which PE ratings capture only indirectly through evaluator preferences for certain measures.

\end{appendices}

\end{document}